%% file: main.tex
\documentclass[letterpaper, 10 pt, conference]{ieeeconf}  

\IEEEoverridecommandlockouts                              

\usepackage{graphicx} 
\usepackage[export]{adjustbox}
\usepackage{amsmath} 
\usepackage{subcaption}
\usepackage{xcolor}

\usepackage{comment}
\usepackage[skip=2pt]{caption}
\usepackage{url}

\newcommand\TODO[1]{\textbf{\textcolor{red}{#1}}}

\title{\LARGE \bf
PATHoBot: A Robot for Glasshouse Crop Phenotyping and Intervention
}

\author{Claus Smitt$^{1}$, Michael Halstead$^{1}$, Tobias Zaenker$^{1}$, Maren Bennewitz$^{1}$ and Chris McCool$^{1}$
\thanks{$^{1}$The authors are with the University of Bonn, Germany.
        {\tt\small \{csmitt, michael.halstead, tzaenker, cmccool\}@uni-bonn.de,} 	\tt\small maren@cs.uni-bonn.de}
}

\begin{document}

\maketitle
\thispagestyle{empty}
\pagestyle{empty}

\begin{abstract}

We present PATHoBot an autonomous crop surveying and intervention robot for glasshouse environments. 
The aim of this platform is to autonomously gather high quality data and also estimate key phenotypic parameters.
To achieve this we retro-fit an off-the-shelf pipe-rail trolley with an array of multi-modal cameras, navigation sensors and a robotic arm for close surveying tasks and intervention.
In this paper we describe PATHoBot design choices made to ensure proper operation in a commercial glasshouse environment.
As a surveying platform we collect a number of datasets which include both sweet pepper and tomatoes.
We show how PATHoBot enables novel surveillance approaches by first improving our previous work on fruit counting by incorporating wheel odometry and depth information.
We find that by introducing re-projection and depth information we are able to achieve an absolute improvement of $20$ points over the baseline technique in an ``in the wild'' situation.
Finally, we present a 3D mapping case study, further showcasing PATHoBot's crop surveying capabilities.

\end{abstract}


\section{Introduction}
\label{sec:indro}

\input{contribs/introduction}

\section{Related Work}
\label{sec:related}

\input{contribs/related}

\section{Platform Design}
\label{sec:platform}

\input{contribs/platformSoftware}

\section{Platform surveillance enabled approaches}
\label{sec:expsetup}
\input{contribs/experimentalsetup}

\section{Results}
\label{sec:results}

\input{contribs/Results}


\section{Conclusion}

In this paper we introduced PATHoBot, an automated platform for surveying crops, capable of operating in a commercial protected cropping environment.
This robot has an array of localization and multi-modal sensors which produce rich data that can be employed by advanced computer vision and machine learning techniques to estimate key phenotyping indices.
An example of method enabled by PATHoBot was through its use for crop counting of sweet peppers ``in the wild''.
We were able to leverage existing annotated data to train models in one domain and utilise them on different cultivar (even with different fruit color). 
We then enhanced our tracking-via-segmentation approach, outperforming the baseline on the same data by approximately 20 points, while also achieving a considerably improved $R^2$ score. 
This improvement in fruit counting was only possibly due to the information captured by the platform.
We also demonstrated PATHoBot's on-the-fly scanning ability by producing $3D$ visualisations offline from the camera array data combined with motion information.

To the best of our knowledge this is the first robot capable of surveying a whole crop in a commercial glasshouse, generating suitable data for autonomous phenotyping and crop mapping.
We believe this platform helps to simplify laborious phenotyping processes, allowing for more efficient and higher quality crop production.

\section*{Acknowledgements}

This work was partialy funded by the Deutsche Forschungsgemeinschaft (DFG, German Research Foundation) under Germany’s Excellence Strategy - EXC 2070 – 390732324 and partially funded by the TRA Sustainable Futures (University of Bonn) as part of the Excellence Strategy of the federal and state governments.

\bibliography{references}
\bibliographystyle{IEEEtran}


\end{document}

%% file: contribs/introduction.tex

Agricultural robotics is a rapidly moving research field due to advances in computer vision and machine learning, and increased agricultural demand.
Research into horticultural automation has been exploring the potential to automate the monitoring of fruit~\cite{sa2016deepfruits, halstead2018fruit, polder2012phenobot} to estimate phenotypic parameters such as yield and ripeness.
Harvesting in ad-hoc cropping systems~\cite{Lehnert20_1}, and commercial glasshouses~\cite{arad2020development} has also been investigated, these techniques generally use mobile platforms and robot arms capable of locally sensing, mapping, and cutting fruits.
However, there is still a considerable gap between farming requirements and available technology, specifically in terms of sensing capabilities to tackle these challenging and dynamic environments. 

\def \figTrolleyHeight {55mm}
\begin{figure}[t!]
	\centering
	\begin{adjustbox}{width=\linewidth}
    	\includegraphics[height=\figTrolleyHeight]{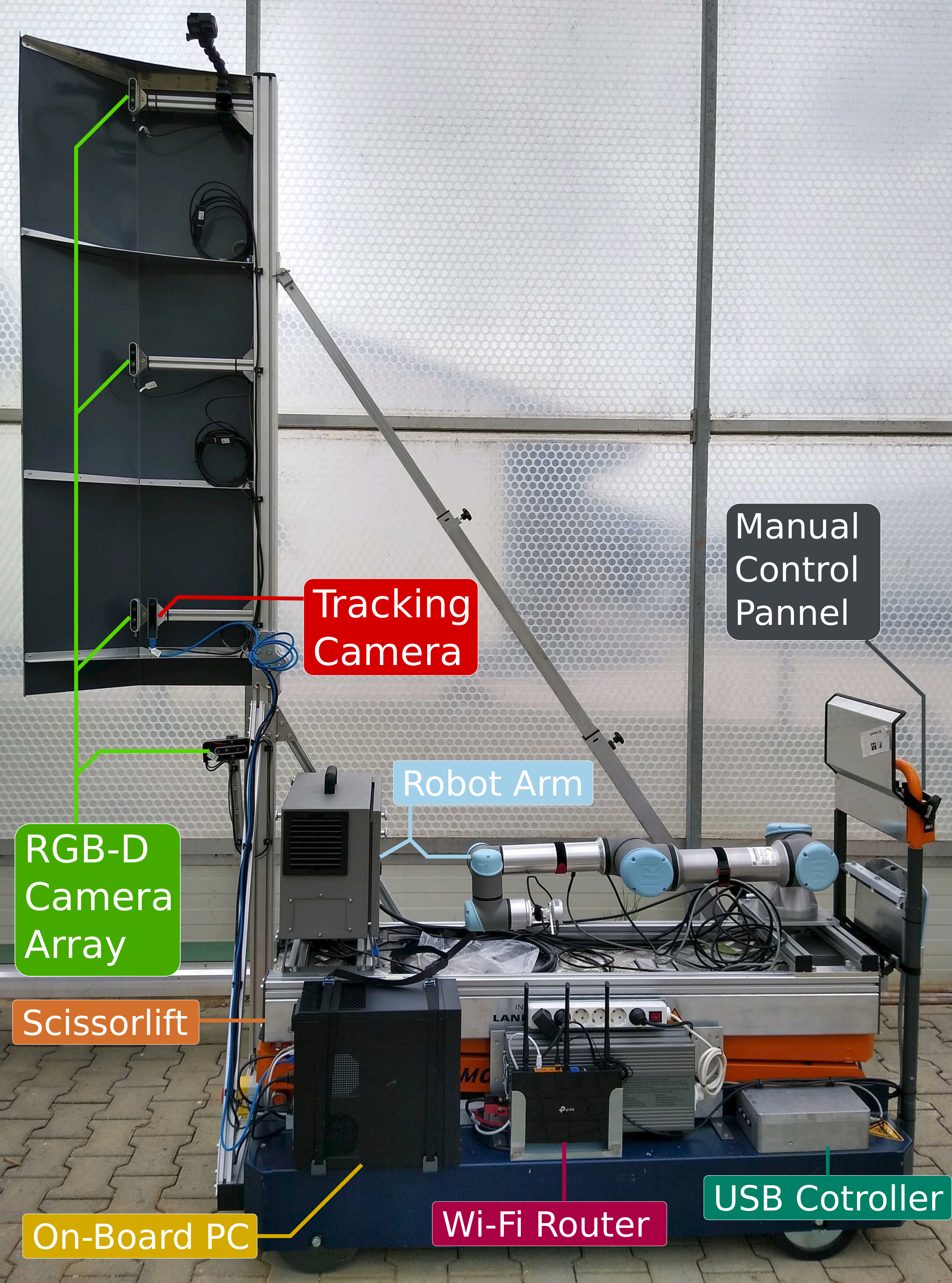}
		\includegraphics[height=\figTrolleyHeight]{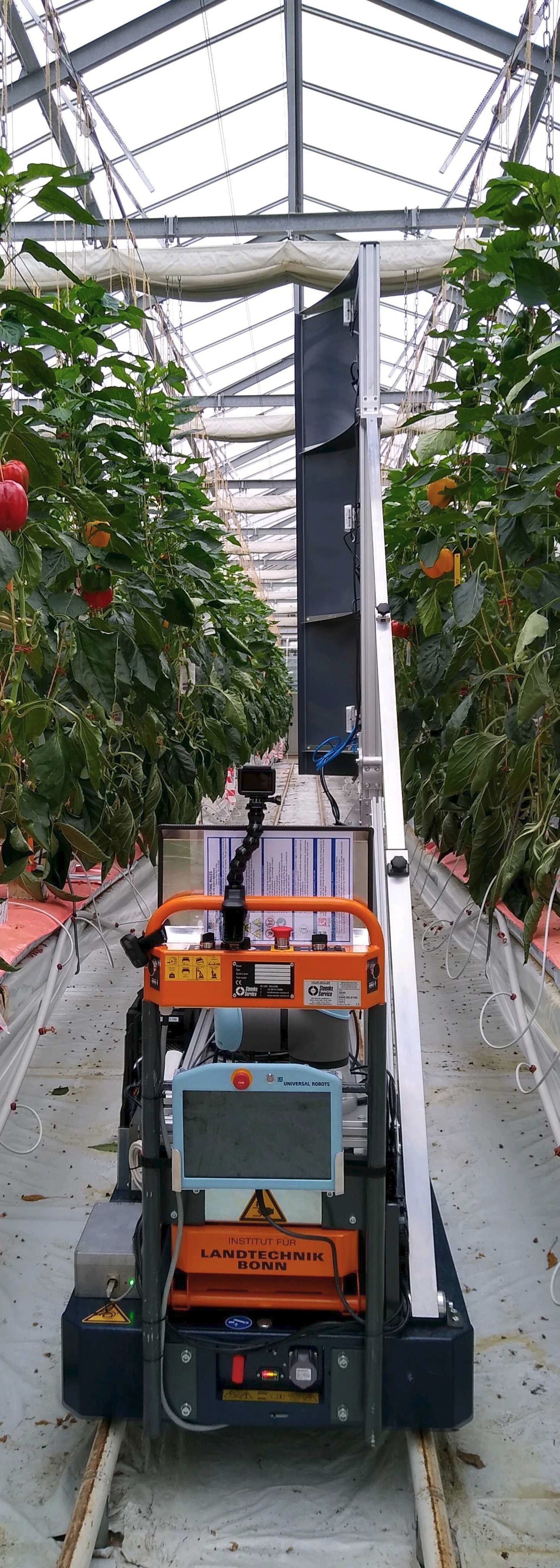}
	\end{adjustbox}
	\caption{Overview of PATHoBot with sensors, arm and other components highlighted and operation in the glasshouse.}
    \vspace{-20pt}
    \label{fig:trolleyComponents}
\end{figure}

This paper introduces PATHoBot (\textbf{P}henotyping \textbf{A}utonomous \textbf{T}rolley for \textbf{Ho}rticulture), an automated robotic platform capable of operating in protected cropping environments. 
The system captures color (RGB), depth (D), and near-infrared (NIR) data of the cropping system using a custom camera array.
Platform motion information is also obtained using encoders and a tracking camera.
This information is suitable to perform in-situ surveillance and to estimate phenotypic indices.
The platform also carries a robot arm intended for close proximity surveying and interaction. 
To evaluate the viability of PATHoBot we capture two species of fruit in a commercial glasshouse setting: sweet pepper and tomato.
We demonstrate the utility of PATHoBot by presenting two surveillance tasks.

First, we use the platform to propose an improvement on previous work on automated fruit counting~\cite{halstead2018fruit}, by leveraging information made available by this robot.
We employ existing models trained in a similar domain on different cultivar (and of different color) to perform ``in the wild'' tracking-via-segmentation to count the number of fruit seen by the platform. 
We investigate the use of the depth and motion information from the platform for both mask re-projection and distance based filtering.
This exemplifies how incorporating spatio-temporal information from this platform can be used to enhance existing techniques.

Second, we demonstrate our system's ability of surveying a whole crop-row on-the-fly by producing $3D$ visualisations from the data gathered offline. 
We achieve this by concatenating point-clouds, from our RGB-D sensor array, and using odometry data to show them in a common reference frame as a single map.

This leads to the main contributions of this work being:
\begin{itemize}
	\item A ROS enabled, crop surveying platform running in a commercial glasshouse environment;
	\item A sensing approach to scan entire crop rows on-the-fly; 
	\item A multi-modal vision-based fruit counting approach exploiting spatial-temporal information made available by this platform;
	\item 3D maps generated from the gathered datasets, enabled by our on-the-fly surveying method. 
\end{itemize}
We present the related work in Sec.~\ref{sec:related}, and the cropping system and our platform design consideration in Sec.~\ref{sec:platform}.
Sec.~\ref{sec:expsetup} presents our fruit counting and mapping experiments enabled by this platform and results are shown in Sec~\ref{sec:results}.

%% file: contribs/related.tex
Multiple research groups have tackled the task of automating crop production in glasshouses.
Belforte et al.~\cite{Belforte2006} built a custom 3 DOF robot arm to show the potential use of robots in glasshouse environments, as a proof of concept for a number of tasks through predefined motions. 


Polder et al.~\cite{polder2012phenobot} retro-fitted a spraying platform into a phenotyping robot for commercial glasshouses by allowing it to drive on the heating pipe-rails. 
It carried a single multi-focus plenoptic camera to produce registered RGB-D images, successfully aiding color-based crop detection by depth filtering.
However, the system was very sensitive to illumination changes.
More recently, a system introduced in~\cite{arad2020development} consisted also of a pipe-rail trolley with a scissor-lift platform, which carried an arm to perform local crop semantic segmentation to find a harvesting path.
This system operated autonomously inside crop-rows once snapped to the rails by an operator.
Also employing an arm~\cite{lehnert20183d}, uses a multi-camera array to obtain multiple samples of the crop from different viewpoints.

In 2017 Lehnert et al.~\cite{lehnert2017autonomous} presented a sweet pepper harvesting robot for ad-hoc cropping systems. More recently, Halstead et al.~\cite{halstead2020} acquired a subset of data from field grown sweet pepper, arranged to suit the robotic system.
In both cases sensing capabilities were limited to performing local detection and/or 3D mapping of each crop and left the task of autonomously surveying the entire row as future work.

More recently Kirk et al.~\cite{kirk2020b} tackled strawberry harvesting in poly tunnels with an autonomous navigating Thorvald platform~\cite{grimstad2017thorvald}.
This employed a bottom-mounted multi-view RGB-D camera array to detect and localize fruits for a robot arm to pick them with a specialized gripper.


One of the main drawbacks of mounting a single image sensor on a robot (or arm) is the need to scan crops multiple times from different viewpoints to obtain complete crop coverage; due to the small sensor FOV and space limitations.
We address this by building a sensor array whose combined FOV covers the entire crop, scanning an entire row in a single pass. Similar to~\cite{kirk2020b} but for tall vertical cropping systems.

Crop surveillance data gathered by robots have also enabled modern computer vision and learning based techniques to automate phenotyping processes as outlined below.

\subsection{Fruit segmentation and vision-based phenotyping}
Sa et al. showed in~\cite{sa2016deepfruits} that deep neural network (DNN) approaches produced excellent results for fruit detection by using multi-modal information, namely RGB and NIR images.
Halstead et al.~\cite{halstead2018fruit} worked upon this approach to produce quality and yield estimations of sweet pepper from RGB video sequences, however, their approach does not consider depth information. 
Koirala et al.~\cite{koirala2019deep} compared RCNN and YOLO DNNs for mango detection on-the-fly, from RGB image sequences.

Other researchers explored supervised learning segmentation based approaches such as~\cite{milioto2018real} for crop and weed classification. 
Zabawa et al.~\cite{zabawa2019detection} also tackled this through the challenging task of grape segmentation by biasing a DNN architecture to segment the fruit's edge as a separate class.

Employing depth aware sensors is beneficial not only for crop size estimation, but also for 3D mapping~\cite{lehnert2017autonomous, arad2020development} and as extra information for multi-modal segmentation~\cite{eitel2015multimodal,valada2019self}.
Detection/segmentation and 3d mapping techniques can be combined to generate semantic reconstructions as shown by McCormac et al.~\cite{mccormac2017semanticfusion}.
These techniques surged in recent years thanks to developments based on autonomous driving~\cite{behley2019semantickitti} and have the potential to be transferred to semantic crop mapping applications as well as fruit and organ-wise phenotyping.


%% file: contribs/platformSoftware.tex
\def \figGlashouseHeight {58mm}
\begin{figure}[t!]
    \centering
    
    \begin{subfigure}{0.59\linewidth}
        \centering
        \vspace{2mm}
    	\includegraphics[height=\figGlashouseHeight]{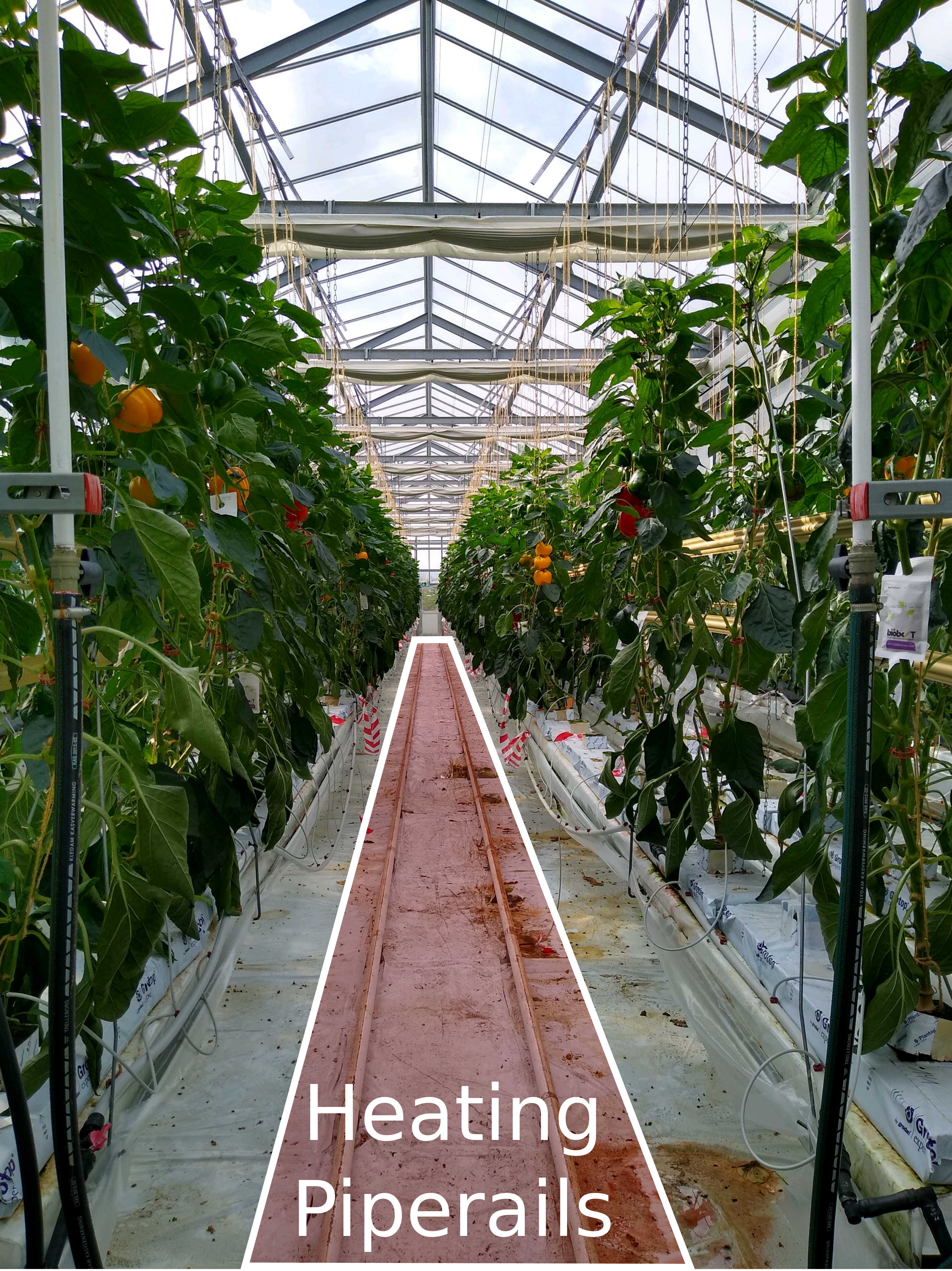}
    	\caption{}
    	\label{fig:chamberRails}
    \end{subfigure}
    \begin{subfigure}{0.39\linewidth}
        \centering
    	\includegraphics[height=\figGlashouseHeight]{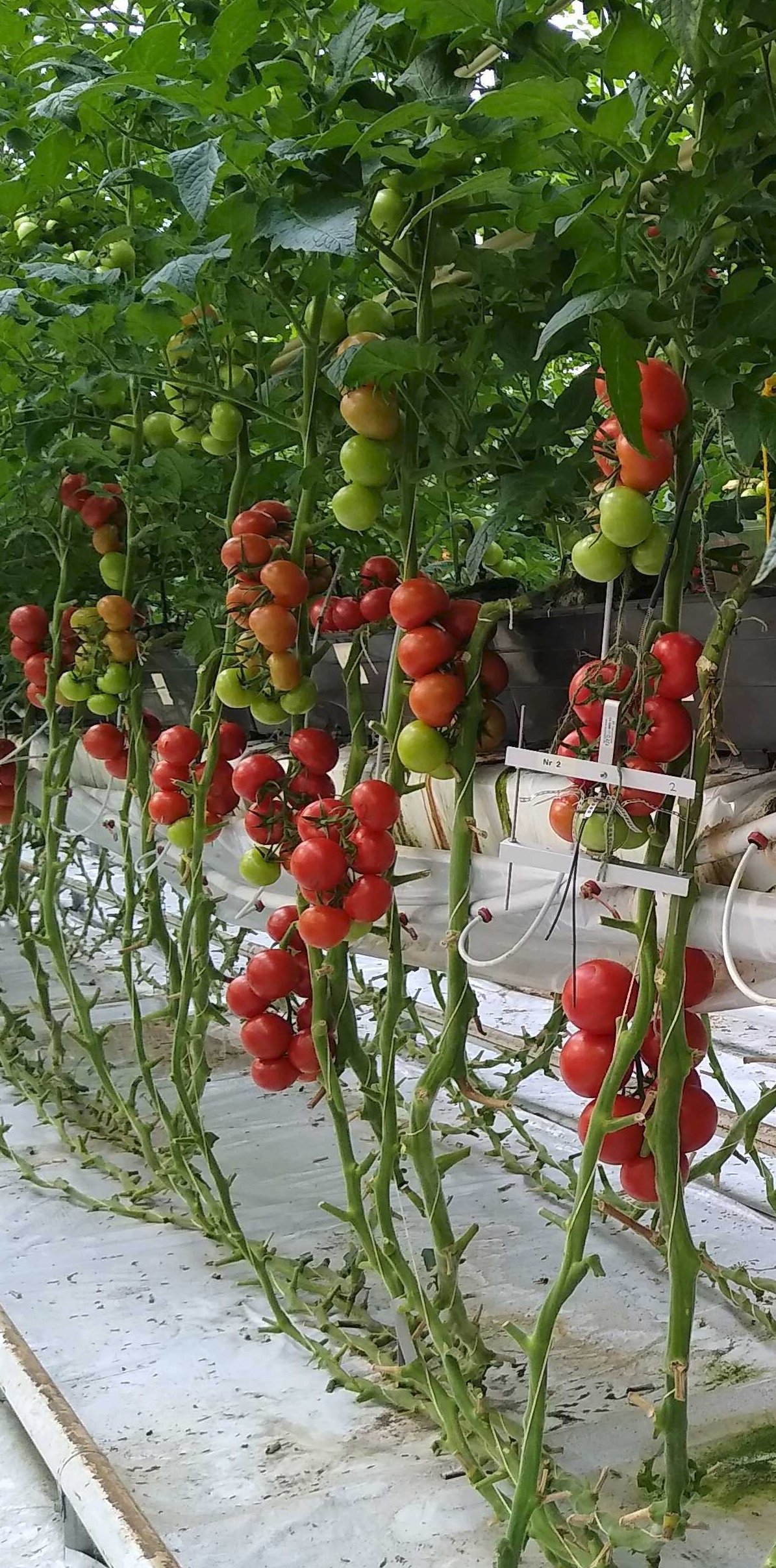}
    	\caption{}
    	\label{fig:tomatoDangle}
    \end{subfigure}
    \vspace{-3pt}
    \caption{a) Sweet Pepper chamber with highlighted pipe-rails; b) tomato close up showing dangling vines}
    \label{fig:glasshouse}
    \vspace{-15pt}
\end{figure}

The PATHoBot platform is designed to operate in commercial glasshouses, leveraging existing glasshouse infrastructure, reducing the barriers for grower adoption.
To outline the decisions behind our design, we first describe the glasshouse infrastructure and then define the design of the platform.

\subsection{Cropping System Environment and Infrastructure}
\label{sec:env}

Our platform is focused on glasshouse crops and makes use of the University of Bonn's commercial glasshouse located at Campus Klein Altendorf (CKA).
It consists of multiple automated chambers capable of self-regulated heating, shading, ventilation and irrigation. 
Heating is achieved using pipe-rails mounted on the ground, which are also used to move machinery, see Fig.~\ref{fig:chamberRails}.
Each chamber consist of 6 rows of hanging substrate trays where the crops are grown and each row is $34m$ long. 
Here, glasshouse crops are trellised and can reach several meters in height.
Two crops, tomato and sweet pepper, are routinely grown in CKA and so are the focus for this platform.

Cultivated species during this study were sweet pepper \textit{Mavera} (yellow) and \textit{Allrounder} (red), and \textit{Lyterno RZ F1} tomatoes. 
Generally, sweet pepper leaves grow larger than the fruit which can occlude them partially or completely. 
A further complication with sweet pepper is the juvenile fruit leaves have a similar green tone, making them difficult to distinguish.
For tomatoes, as long as weather conditions are favorable, the stems will grow continuously.
This is managed at CKA such that the fruit stays low to simplify cropping, even dangling below their substrates (see Fig.~\ref{fig:tomatoDangle}). 
Moreover, the leaves are smaller than sweet pepper ones, making occlusions less frequent, and juvenile tomato fruits have a lighter green tone, making them easier to spot.


At CKA, motorized pipe-rail trolleys, with a manually set height, are employed for crop management tasks (e.g. pruning and harvesting).
This infrastructure is also used to measure phenotypic traits using specialized sensors, these tasks are currently time consuming due to a lack of automation. 
%
One limitation of the CKA glasshouse is the available space to perform turns at the end of each row as there is less than ${\sim}1.5m$ of free space.
Moreover, there are also beams, piping and wires which impose height constraints (${\sim}2m$) for operating in these areas.
From these environment conditions and current operations carried out in CKA we derived the following requirements for our robot:
\begin{itemize}
    \item a platform capable of navigating through rows autonomously, carrying surveying sensors and actuators;
    \item leverage glasshouse infrastructure (e.g. pipe-rails);
    \item carry an array of high quality multi-modal sensors, capable of capturing whole plants on-the-fly, and an extra navigation sensor to enable consistent data fusion;
    \item and carry a robot arm capable of reaching and potentially manipulating all crop. 
\end{itemize}

\subsection{Retrofitted commercial cropping platform}
\label{ssec:retrofit}
PATHoBot is built around an off-the-shelf glasshouse platform which fits within the limited space at the end of the row. 
It has a pneumatic actuated scissor-lift and is capable of carrying loads of up to 455 Kg and lift up to 3m, making it ideal to mount a robot arm. 
%
We added a manually collapsible mast upon which the camera array is mounted.
It remains upright while scanning crops and is collapsed to avoid the infrastructure present between the rows (see Sec.~\ref{sec:sensorArray}).



To enable automated operation, we installed a wheel encoder, which is fed to the platform controller. 
This device also receives serial commands over USB, and actuates both the drive motor and scissor-lift.
Finally, we have mounted a DC-AC sinusoidal converter to power all on-board systems with the platform batteries and enable flexible expansion.

\subsection{Sensor Array Design}
\label{sec:sensorArray}

As shown by~\cite{sa2016deepfruits}, multi-modal information improves crop detection results, in particular fusing RGB and NIR information for sweet pepper detection. 
Furthermore, depth information is useful for fruit size estimation and to distinguish between elements in the scene (e.g. crop and leaves). 

To obtain RGB, NIR, and depth information we chose to use the \textit{Intel RealSense D435i} cameras, as it uses NIR images to estimate stereo depth which can be registered to RGB images.
Due to the abundance of texture in the scene the camera IR projector is not required.


\def \figGlashouseHeight {80mm}
\begin{figure}
	\centering
	\vspace{2mm}
	\begin{subfigure}[t]{.47\linewidth}
		\centering
		\includegraphics[height=\figGlashouseHeight]{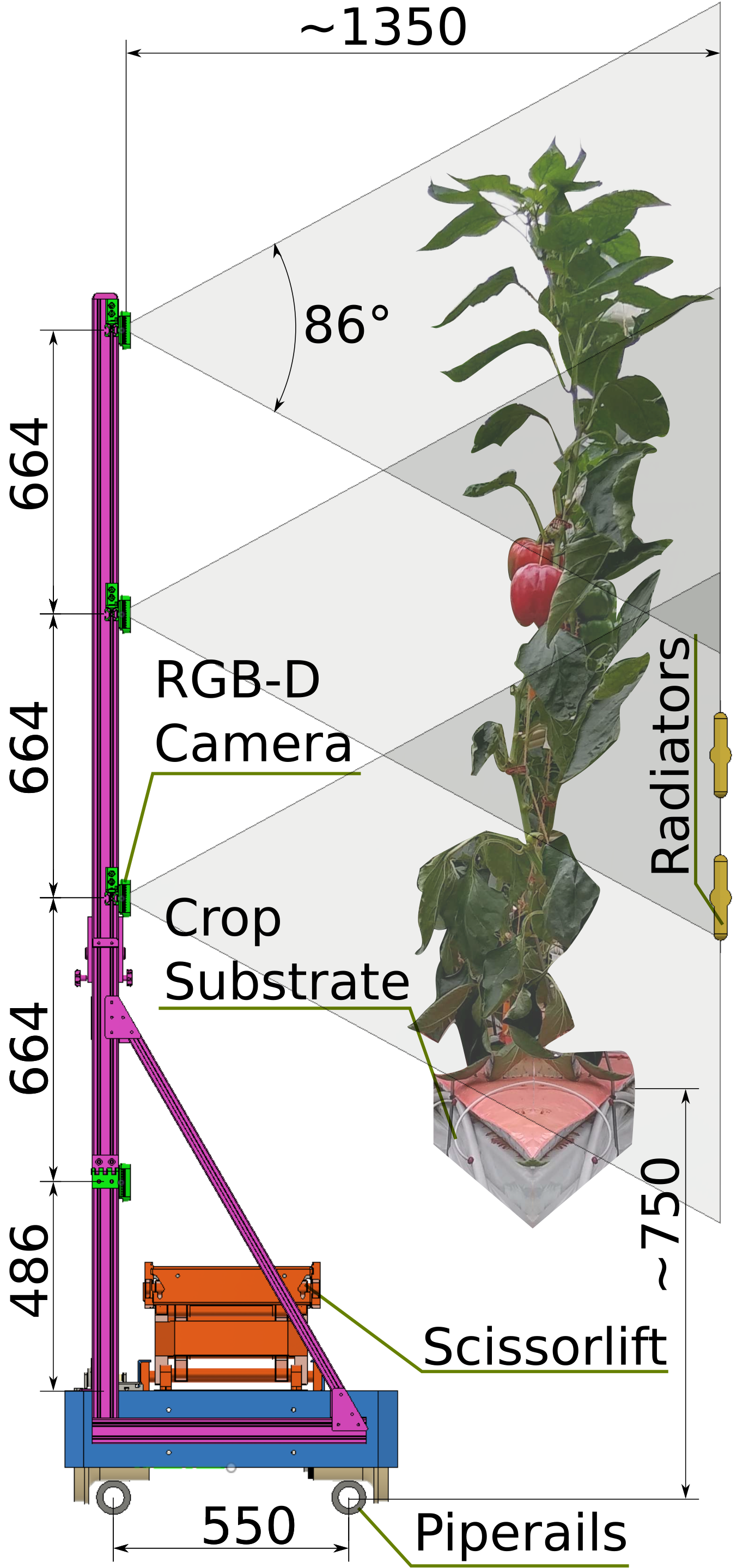}
 		\caption{}
 		\label{fig:FOVpepper}       
	\end{subfigure}
	\begin{subfigure}[t]{.50\linewidth}
		\centering
		\includegraphics[height=68.5mm]{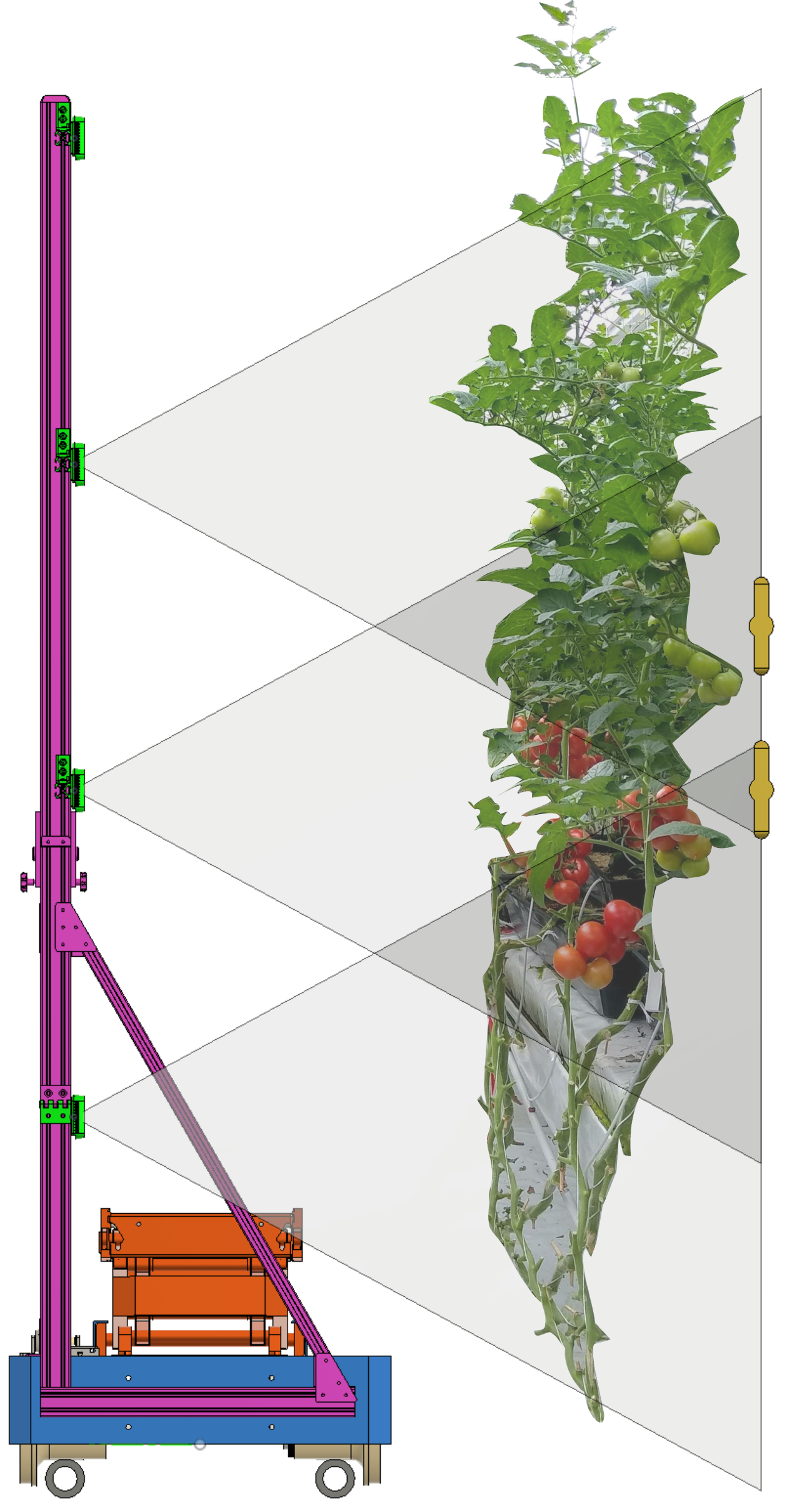}
 		\caption{}
 		\label{fig:FOVtomato}
	\end{subfigure}
	\vspace{-4pt}
	\caption{Crop sensing coverage by our multi RGB-D camera array extended FOV for a) sweet pepper (3 upper cameras) and b) tomato (3 lower cameras to capture vines). }
	\label{fig:FOV}       
	\vspace{-15pt}
\end{figure}

From Fig.~\ref{fig:FOV} we see the general structure of the camera array on the machine, the three upper cameras are used for sweet pepper and the three lower cameras for tomatoes (due to the low lying fruit).
We ensure that a FOV superposition between each camera of approx.~$20\%$ at the crop stem average depth.
Horizontally we space the cameras ${1.35m}$ away from the heating rails, creating greater scene coverage.

As stated in Sec.~\ref{ssec:retrofit} we fit PATHoBot with a collapsible mast to bypass obstacles, only the top three sensors are mounted in this mast. 
Finally, we mount a plastic shield to the mast to prevent damaging crops, and avoid lens occlusions from leaves.
We also mounted an \textit{Intel Realsense T265} tracking camera (See Fig.~\ref{fig:trolleyComponents}), which provides 6DoF pose estimations on-chip. 
This camera will be used to bootstrap consistent 3D mapping and serves as a baseline for any other SLAM algorithms that might be tested in the future.

%

\subsection{On-Board Robot Arm}

A UR5e robot arm from Universal Robots is mounted on the scissor-lift platform.
It has a reach of $850mm$ and can be equipped with payloads of up to 5kg.
Combined with the scissor-lift maximum height of $3m$ the arm is able to reach the entire crop. 
Currently, an Intel RealSense L515 LiDAR camera is used as sensor on the tip of the arm.
The camera is used in combination with a viewpoint planning approach to enable more complete 3D models of the plants by maneuvering the sensor around in order to avoid occlusions by leaves, as described in~\cite{zaenker2020viewpoint}.

To complement the existing camera, other sensors could also be used in the future that require close proximity to parts of the plant.
For example, fluorometers can be used to measure plant physiological indices such as chlorophyll, flavonols, and anthocyanins content in leaves and fruits.
%
Additional intervention based tools could also be mounted to the arm for autonomous interaction.
The arm's control box is also connected to the platforms wired network, and commanded via standard ROS interfaces such as MoveIt!. 


\subsection{Software Infrastructure}
\label{sec:software}

The platform has an on-board computer running ROS.
This computer communicates with the platform's USB controller, exposing its state and receiving commands through the ROS network.
To enable remote operations, we employ an on-board Wi-Fi router to ensure reliable communication with remote stations; however, only non-critical mission data is exchanged through this link for safety reasons.

As explained in Sec.~\ref{sec:sensorArray} only 3 of the cameras will be recording simultaneously.
Each one records RGB, NIR and depth images (at $15Hz$, 1280x720) as well as IMU data (gyro at $400Hz$, and accel at $250Hz$) which amounts to a total bandwidth of approx~$1.8Gb/s$.
We performed extensive testing of multiple hardware configurations (for these 3 cameras) and found that a dedicated PICe card was required to ensure no frame drops. 
To ensure synchronization, the 3 cameras were triggered as described in~\cite{grunnet2018using}.
Furthermore, to deal with variable illumination we ensured that all of the cameras used automatic exposure and white-balance.

The on-board computer is a small form factor desktop PC equipped with a graphics processing unit (GPU). 
To enable running high computational load algorithms on-board, such as fruit detection, 3D mapping and navigation, the system includes an i7-10700K CPU, 32Gb RAM and an NVidia GeForce RTX 2080Ti.






%% file: contribs/experimentalsetup.tex
The following sections describe pehnotyping related applications enabled by PATHoBot sensing capabilities.
We first exploit spatial-temporal information for counting sweet pepper in a row, and finally we perform autonomous crop 3D mapping.


We scan sweet pepper at a speed of approximately $0.2m/s$. 
At this speed, scanning a whole chamber takes approximately $50$mins, making it suitable for frequent surveying.
We capture a number of datasets over a three month period with different separation between captures.
%
We select the final two captured datasets to evaluate our platform enabled surveillance technique. 
These datasets capture RGB, NIR, and depth images along with wheel encoder information as outlined in Section~\ref{sec:software}.

\subsection{Fruit Counting}
\label{sec:setupFruit}

To improve the spatial awareness of our previous work on fruit counting~\cite{halstead2018fruit} we investigate adding two components to the prior IoU tracking strategy, both of which exploit spatial-temporal information captured by this platform. 
First, we re-project detection tracklets for comparison with new detections, creating a more precise estimate of where the fruit should appear in the next image. 
This leads to higher IoU comparisons between the tracklet and associated detections.
Second, we further exploit depth information to filter objects outside of a specific range.
As discussed in Section~\ref{sec:platform}, detecting/segmenting sweet pepper in a commercial glasshouse is a challenging task due to occlusion, juvenile sweet pepper, similar appearance to leaves, and strong illumination variation.
We aim to count all of the sweet pepper in a row recorded by the RGB-D camera array, using only the second camera from the top.
This involves a multi-staged process including instance segmentation utilising Mask-RCNN~\cite{he2017mask} and tracking-via-segmentation based on~\cite{halstead2018fruit}.


Our Mask-RCNN network is trained on the recently released data from~\cite{halstead2020}.
While the data was captured in a similar environment (glasshouse) the training data in~\cite{halstead2020} contains black and red sweet pepper species, while the evaluation data here contains yellow, and red (in all its development stages, green to ripe), and stronger illumination changes.
This variation imposes a domain shift between the datasets creating an ``in the wild'' situation.

The counting component of this experiment is a modified version of the tracking-via-detection approach in~\cite{halstead2018fruit} to incorporate segmentation masks. 
This technique tracks a sweet pepper through a video sequence to count it only once by associating new detections with active tracklets.
A number of hyper-parameters are employed to accurately calculate the association of detection and tracklets but also ascertain if a tracklet meets the conditions of being a legitimate fruit.



To achieve re-projection we exploit the RGB-D information captured by the platform and employ the \textit{Intel Realsense}'s API to register the depth to the color and the factory calibrated intrinsic parameters for a pinhole camera model as described in~\cite{hartley2003multiple}.
To re-project a detection mask from (image) frame $i$ to $j$, we can compute the camera homogeneous transform $Tc_{ij}$ by using the wheel odometry transform between them $Te_{ij}$. We also account for the camera extrinsics to the encoders $T_{ec}$ (based on the platform's CAD model) as follows, 
\begin{equation}
    Tc_{ij} = T_{ec}^{-1} Te_{ij} T_{ec}.
\end{equation}
Given a binary detection mask in frame $i$ composed by pixels coordinates $M_i = \{\mathbf{m}_{i1},\mathbf{m}_{i2},\dots,\mathbf{m}_{iN}\}$ the mask points can be re-projected into frame $j$,
\begin{equation}
    \mathbf{m}_{jk} = \pi(Tc_{ij}(\pi^{-1}(\mathbf{m}_{ik},d_{m_{ik}})))  \ / \ k \ \epsilon \  [1,\dots, N],
\end{equation}
where $\pi(.)$ is the camera projection function, $d_{m_.}$ is each mask coordinate's depth and $T(.)$ applies a homogeneous transform to a 3D point.
This provides a more up-to-date spatial mask of the tracklet for comparison to new detections and also provides the potential to track fruit through small occlusions.


In the final evaluation, along with re-projection, we perform depth-based detection filtering, similar to~\cite{polder2014depth}. 
For the $i$-th frame (image) we obtain $K$ detections consisting of masks $\mathcal{M}^{i} = \left[M_{i1}, M_{i2}, \dots, M_{iK} \right]$.
We only retain the detections if a sufficient proportion of their depth values are within the closest row, $d_{crop}=[0.2,1.4]m$,
\begin{equation}
    \frac{\#\{d_{m_{ik}}\ \epsilon \ d_{crop}\}}{\#\{\mathbf{m}_{ik}\}} > \tau_{dpt},
    \label{ec:depthTh}
\end{equation}
where $\tau_{dpt}$ was empirically derived, $\#\{\mathbf{m}_{ik}\}$ is the total number of values in the mask $M_{ik}$, and  $\#\{d_{m_{ik}}\ \epsilon \ d_{crop}\}$ is the number of depth values within the defined range $d_{crop}$. 

For our evaluation we will utilise three different tracking set ups, all using the same Mask-RCNN models for mask propagation.
The initial experiment uses a baseline (\textbf{bl}) tracker built in a similar method to~\cite{halstead2018fruit}. 
In the next two evaluations we exploit spatial-temporal information by utilising re-projection only (\textbf{rp}), and re-projection plus depth filtering in the final (\textbf{df}).
To evaluate all approaches we use normalised error $NE$ of total fruit count in the current row,
\begin{eqnarray}
  NE = \frac{|GT - pred|}{GT},
\end{eqnarray}
where $GT$ is the manually annotated visual fruit count from RGB images in each row, done by a single person, and $pred$ represents the predicted fruit count from the tracking-via-segmentation approach.
We also utilise the mean error of an experiment to display the overall performance at each IoU value.
Finally, we compute the $R^2$ cofficient of detrmination to measure the correlation between the $GT$ and the prediction.

\subsection{Crop Sensing and Mapping}
\label{sec:mappingSetup}

We also outline a method for autonomous crop mapping, enabled by PATHoBot's on-the-fly surveying capabilities.
We first synchronize all camera feeds, utilising the built in triggering hardware (Sec.~\ref{sec:software}).
Then we employ \textit{Intel Realsense}'s API to generate colored 3D point-clouds from registered RGB-D images, while keeping track of the platform state and frames with a robot model (see Fig.~\ref{fig:mapRow}). 
Each camera's pointcloud is referenced to that robot model, producing spatially referenced point-clouds.
Furthermore, we filter points outside the surveyed crop-row by keeping those with depth $d\ \epsilon \ [0.2,1.4]m$.
Since this simple experiment concatenates frame-wise point-clouds on a single map, we skip ${\sim}60$ frames to get minimal superposition between point-clouds at the radiators depth (see Fig.~\ref{fig:FOV}), yielding a less dense visualization. This simple pipeline is computationally-inexpensive and can run real-time on a single core of a modern PC. 


%% file: contribs/Results.tex
The following section details the results of our experiments performed on the last two datasets gathered by PATHoBot at CKA commercial glasshouse.
These evaluations outline the versitility of PATHoBot for various agricultural robotic based tasks.

\begin{figure}[t]
    \centering
    \vspace{2mm}
    \includegraphics[width=\linewidth]{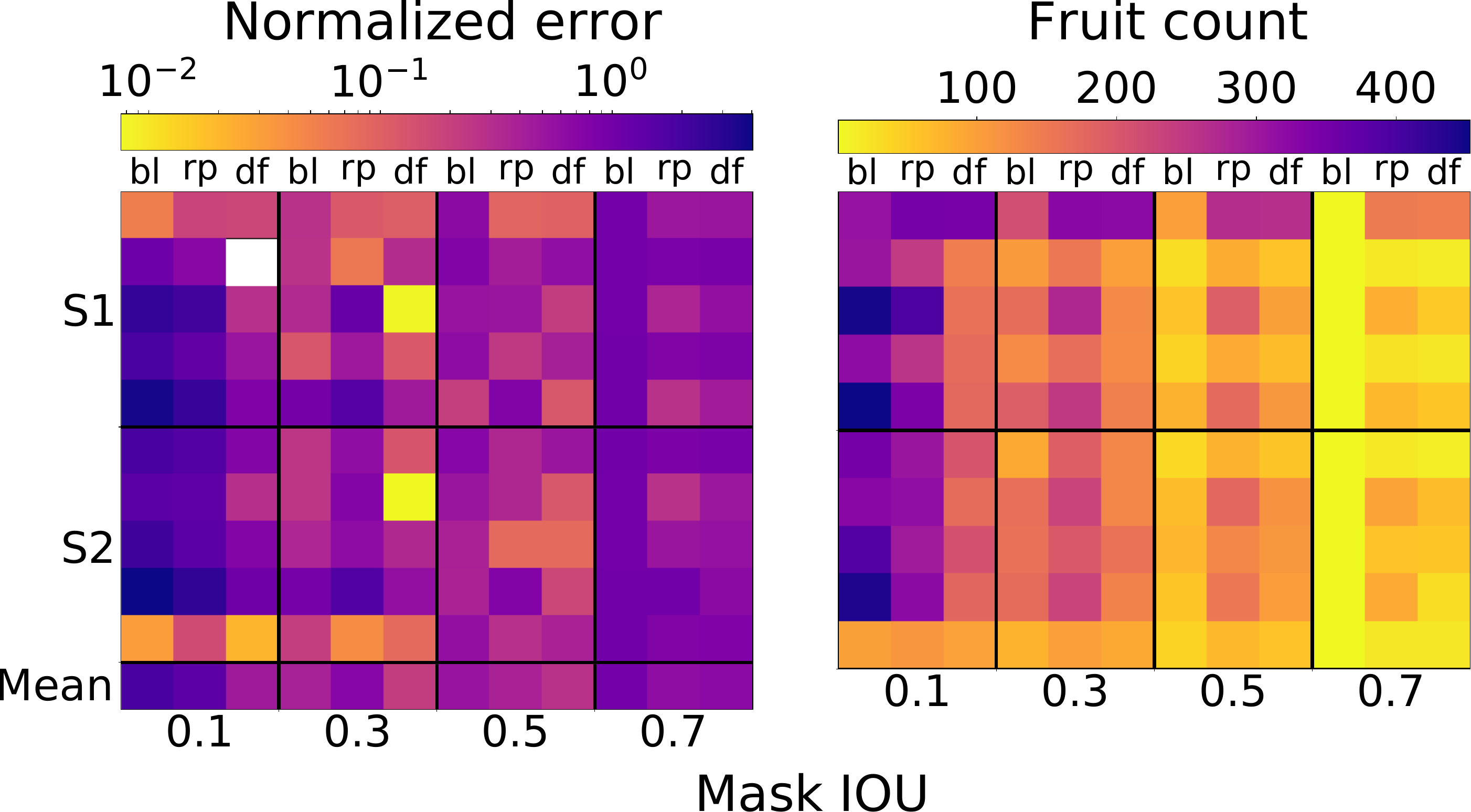}
    \caption{Fruit count experiments results for different IoUs and baseline system (bl); re-projection approach (rp); re-projection plus depth filtering (df). Each row is the result of processing a crop-rows scan in 2 surveying sessions (S1, S2). White cells represent 0 values.}
    \label{fig:cntMetrics}
    \vspace{-15pt}
\end{figure}

\def \figMapsHeight {38mm}
\begin{figure*}[ht!]
    \centering
    \vspace{2mm}
    \begin{subfigure}{0.14\textwidth}
        \centering
    	\includegraphics[height=\figMapsHeight]{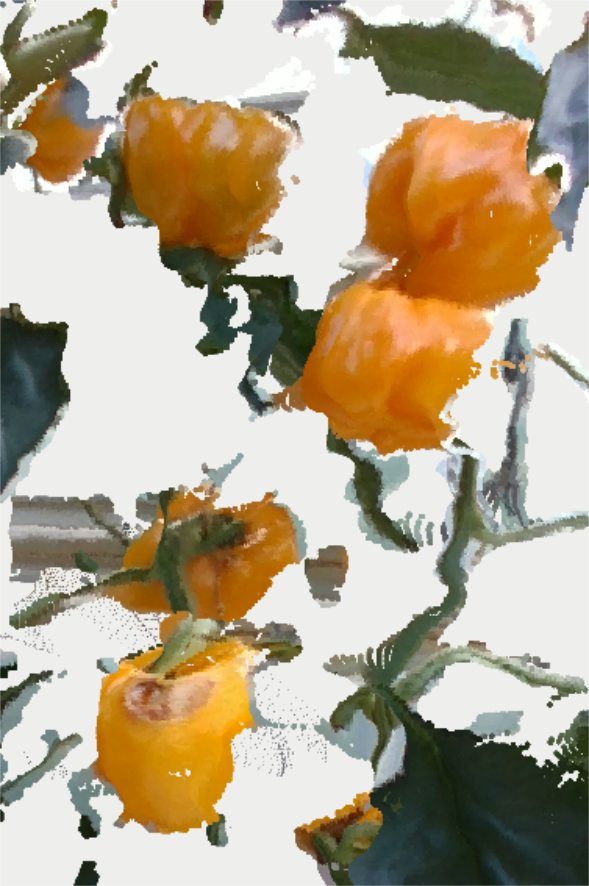}
    	\caption{}
    	\label{fig:closepuRed}
    \end{subfigure}
    \begin{subfigure}{0.14\textwidth}
        \centering
    	\includegraphics[height=\figMapsHeight]{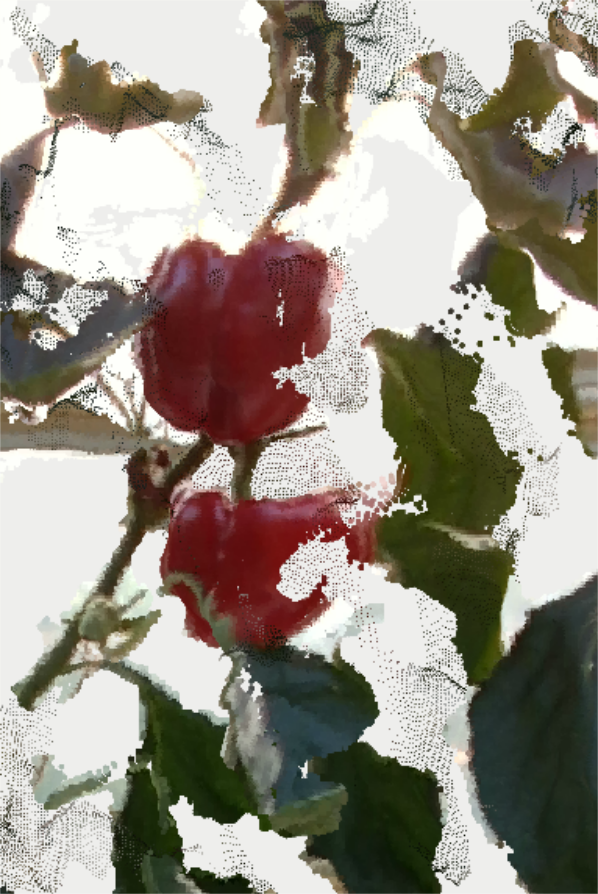}
    	\caption{}
    	\label{fig:closepuBlend}
    \end{subfigure}
        \begin{subfigure}{0.69\textwidth}
        \centering
    	\includegraphics[height=\figMapsHeight]{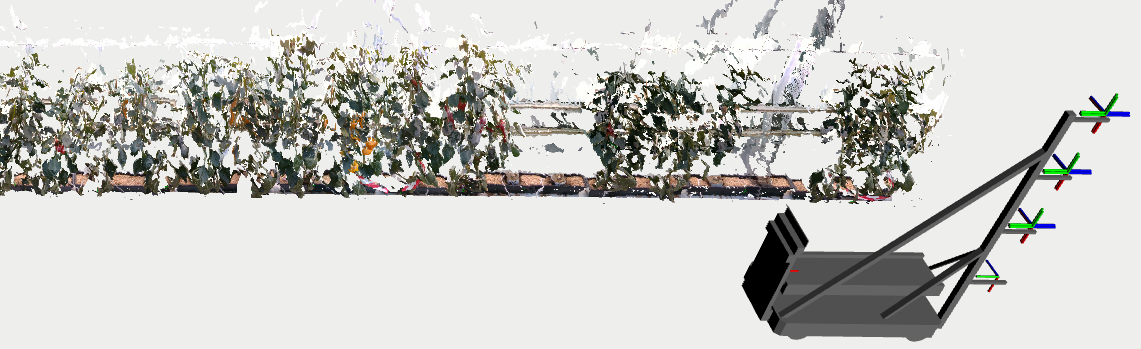}
    	\caption{}
    	\label{fig:mapRow}
    \end{subfigure}
    \caption{Sweet pepper mapping results showing a) multiple mapped fruits; b) fruit blending with leafs and holes due to depth interpolation and occlusions respectively; c) section of a sweet pepper row 3D map also showing the robot state.}
    \label{fig:my_label}
    \vspace{-15pt}
\end{figure*}

\subsection{Fruit Counting}
Based on the techniques and data outlined in Sec.~\ref{sec:setupFruit} we evaluate a Mask-RCNN instance based segmentation network, due to its good performance on ``in the wild'' scenarios \cite{halstead2020fruit}.
We also implement a version of the IoU-based fruit tracker in~\cite{halstead2018fruit}, to count fruit instances in a scene.

The tracking-via-detection approach outlined in~\cite{halstead2018fruit} exploits the accuracy of the detection routine (Faster-RCNN in this case) for track objects in consecutive frames.
We perform the same task, but replace bounding-box tracking with instance-based segmentation tracking (using Mask-RCNN). 
Initially, we ensure sweet pepper are completely visible in the frame, creating a ``tracking region''.
Segmentation outputs within this ``tracking region'' are either stored as new tracklets or assigned to an existing tracklet using an IoU metric.
To remove spurious detections, only tracklets with at least 10 assigned frames (using the IoU metric) are retained; more details can be found in~\cite{halstead2018fruit}.
Finally, to ensure we filter detections from other rows, we set a depth filter threshold (see eq.~\ref{ec:depthTh}) of $0.5$m based on visually inspecting results.

The segmentation stage takes $0.07$s (approx. $14$fps) to run on a robot computer, making it near real-time. However, the tracking stage takes $0.96$s since it is a proof-of-concept and its optimization is out of the scope of this work.

Fig.~\ref{fig:cntMetrics} provides an overview of the performance of all the experiments, run on $120s$ long excerpts ($1800$ frames) of each sequence.
Generally, we see improved accuracy through the three systems where \textbf{bl} performs the worst and the filtered re-projection (\textbf{df}) outperforms the other two.
This improved performance can be seen in Fig.~\ref{fig:cntMetrics} and we attribute this improved performance to both filtering and re-projection; only relevant fruit are detected and tracklet IoU is improved.
This can be seen by investigating the performance of re-projection (\textbf{rp}) on its own.
In the Fruit Count plot of Fig.~\ref{fig:cntMetrics} it can be seen that \textbf{rp} has a consistently higher estimate of the number of fruit (across all IoUs) when compared to \textbf{df}.
We believe that this is because \textbf{rp} tracks sweet pepper from other rows, inflating the overall count of this system.

\begin{figure}
    \centering
    \includegraphics[width=0.325\columnwidth]{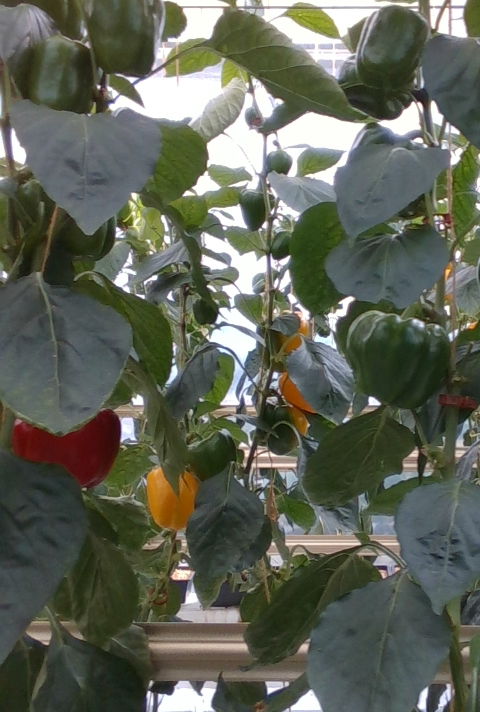}
    \includegraphics[width=0.325\columnwidth]{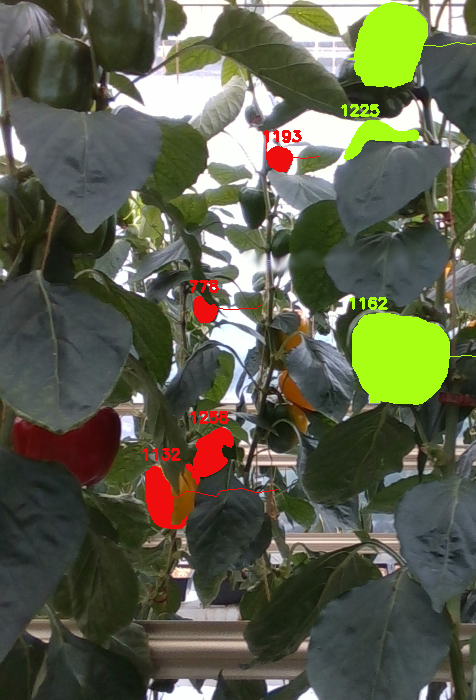}
    \includegraphics[width=0.325\columnwidth]{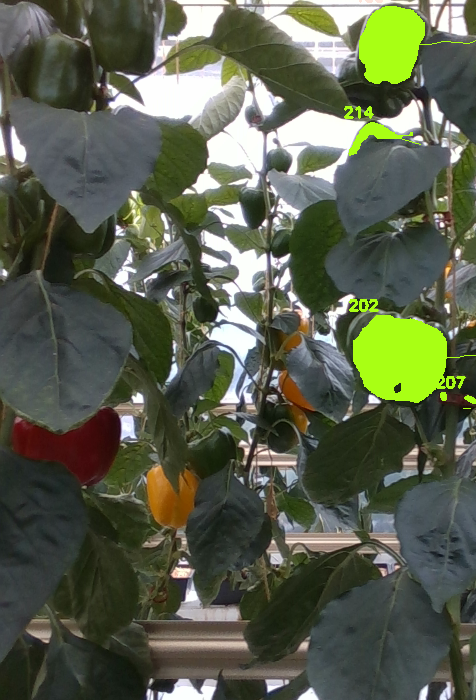} 
    \\
    \caption{Benefits of including the depth filtering. From left to right: raw image, unfiltered segmentation and depth filtered. Sweet peppers that don't belong to the current row (red) are filtered out, leaving the closer ones (green) to be tracked.}
    \label{fig:depthFilt}
    \vspace{-15pt}
\end{figure}


\begin{table}[b]
    \vspace{-15pt}
    \centering
    \caption{Results of the fruit counting for IoU=$0.3$.}
    \label{tab:exampleresults}
    \resizebox{\columnwidth}{!}{%
    \begin{tabular}{|l|c|c|}
    \hline
    & Normalized Error & $R^2$ \\
    \hline\hline
    No Depth (\textbf{bl}) & $0.39\pm{0.28}$ & $0.22$\\
    Reprojection Only (\textbf{rp}) & $0.71\pm{0.56}$ & $0.36$ \\
    Reprojection + Filtering (\textbf{df}) & $0.21\pm{0.19}$ & $0.80$ \\
    \hline
    \end{tabular}
    }
\end{table}

The limitation of the \textbf{rp} technique is highlighted in Fig.~\ref{fig:depthFilt}.
It can be seen that multiple tracks exist in the background rows, which should not be counted.
By introducing a depth-based filtering (\textbf{df}) we are able to remove these unwanted tracks.
This increases the performance of both the mean normalized error and $R^{2}$.
Also, it is noteworthy that the segmentation algorithm detects yellow sweet pepper in an ``in the wild'' scenario, despite no exposure to it during training.

The \textbf{df} technique is the best performing system, at a detection IoU of $0.3$ as seen in Tab.~\ref{tab:exampleresults}.
In terms of overall performance it achieves a mean normalized error of $0.21$.
This is an absolute improvement of $0.18$ over \textbf{bl} which is the next best performing technique.
The \textbf{df} technique also achieves an $R^2$ value of $0.80$ showing a strong relationship between the fruit tracking predictions and the ground truth. 
Thus, in a practical application a simple linear model could be used to improve the mean normalized error of fruit count.
\subsection{Crop Sensing and Mapping}


The mapping pipeline described in~\ref{sec:mappingSetup} is able to yield offline 3D visualizations of each crop-row.
Fig.~\ref{fig:mapRow} outlines the entire height of a single section of one row. 
Portions of this map highlighting the positive and negative attributes are provided in Fig.~\ref{fig:closepuRed} and Fig.~\ref{fig:closepuBlend} respectively.
In the positive example clear dense $3D$ maps of the fruit are obtained.
While the negative example displays promising results, there exists blending between the leaf stem and the peppers surface.
Moreover, un-mapped patches occur due to occlusions.



Despite the simplistic nature of this mapping approach, visualizations should be dense and detailed enough to be used for phenotyping tasks. However, one of the limitations of using point-clouds from a single depth measurement is that they tend to by noisy and graphically intense.
This raises the need for fusing multiple depth measurements in a single map.  Not only to reduce the map noise, but also to associate similar points and produce memory efficient reconstructions. 
Fusing all available information sources (odometry, tracking camera pose, depth images and IMU data) in a SLAM system could produce more precise and consistent 3D maps.


%
Combining such maps with fruit segmentation algorithms should yield rich 3D semantic crop maps suitable for autonomous phenotyping.
This can be further enhanced through $4D$ registration~\cite{magistri2020iros} techniques, accounting for crop growth and enable multi-session crop semantic mapping.

%% file: main.bbl
\begin{thebibliography}{10}
\providecommand{\url}[1]{#1}
\csname url@samestyle\endcsname
\providecommand{\newblock}{\relax}
\providecommand{\bibinfo}[2]{#2}
\providecommand{\BIBentrySTDinterwordspacing}{\spaceskip=0pt\relax}
\providecommand{\BIBentryALTinterwordstretchfactor}{4}
\providecommand{\BIBentryALTinterwordspacing}{\spaceskip=\fontdimen2\font plus
\BIBentryALTinterwordstretchfactor\fontdimen3\font minus
  \fontdimen4\font\relax}
\providecommand{\BIBforeignlanguage}[2]{{%
\expandafter\ifx\csname l@#1\endcsname\relax
\typeout{** WARNING: IEEEtran.bst: No hyphenation pattern has been}%
\typeout{** loaded for the language `#1'. Using the pattern for}%
\typeout{** the default language instead.}%
\else
\language=\csname l@#1\endcsname
\fi
#2}}
\providecommand{\BIBdecl}{\relax}
\BIBdecl

\bibitem{sa2016deepfruits}
I.~Sa, Z.~Ge, F.~Dayoub, B.~Upcroft, T.~Perez, and C.~McCool, ``Deepfruits: A
  fruit detection system using deep neural networks,'' \emph{Sensors}, vol.~16,
  no.~8, p. 1222, 2016.

\bibitem{halstead2018fruit}
M.~Halstead, C.~McCool, S.~Denman, T.~Perez, and C.~Fookes, ``Fruit quantity
  and ripeness estimation using a robotic vision system,'' \emph{IEEE Robotics
  and Automation Letters}, vol.~3, no.~4, pp. 2995--3002, 2018.

\bibitem{polder2012phenobot}
G.~Polder, D.~Lensink, and B.~Veldhuisen, ``Phenobot-a robot system for
  phenotyping large tomato plants in the greenhouse using a 3d light field
  camera,'' \emph{Unpublished lecture}, 2013.

\bibitem{Lehnert20_1}
C.~Lehnert, C.~McCool, I.~Sa, and T.~Perez, ``Performance improvements of a
  sweet pepper harvesting robot in protected cropping environments,''
  \emph{Journal of Field Robotics}, vol.~37, pp. 1197--1223, 2020.

\bibitem{arad2020development}
B.~Arad, J.~Balendonck, R.~Barth, O.~Ben-Shahar, Y.~Edan, T.~Hellstr{\"o}m,
  J.~Hemming, P.~Kurtser, O.~Ringdahl, T.~Tielen \emph{et~al.}, ``Development
  of a sweet pepper harvesting robot,'' \emph{Journal of Field Robotics}, 2020.

\bibitem{Belforte2006}
G.~Belforte, R.~Deboli, P.~Gay, P.~Piccarolo, and D.~{Ricauda Aimonino},
  ``{Robot Design and Testing for Greenhouse Applications},'' \emph{Biosystems
  Engineering}, vol.~95, no.~3, pp. 309--321, 2006.

\bibitem{lehnert20183d}
C.~{Lehnert}, D.~{Tsai}, A.~{Eriksson}, and C.~{McCool}, ``3d move to see:
  Multi-perspective visual servoing towards the next best view within
  unstructured and occluded environments,'' in \emph{2019 IEEE/RSJ
  International Conference on Intelligent Robots and Systems (IROS)}, 2019, pp.
  3890--3897.

\bibitem{lehnert2017autonomous}
C.~Lehnert, A.~English, C.~McCool, A.~W. Tow, and T.~Perez, ``Autonomous sweet
  pepper harvesting for protected cropping systems,'' \emph{IEEE Robotics and
  Automation Letters}, vol.~2, no.~2, pp. 872--879, 2017.

\bibitem{halstead2020}
M.~Halstead, S.~Denman, C.~Fookes, and C.~McCool, ``Fruit detection in the
  wild: The impact of varying conditions and cultivar,'' in \emph{to appear in
  Proceedings of Digital Image Computing: Techniques and Applications (DICTA)},
  2020.

\bibitem{kirk2020b}
R.~Kirk, G.~Cielniak, and M.~Mangan, ``L* a* b* fruits: A rapid and robust
  outdoor fruit detection system combining bio-inspired features with one-stage
  deep learning networks,'' \emph{Sensors}, vol.~20, no.~1, p. 275, 2020.

\bibitem{grimstad2017thorvald}
L.~Grimstad and P.~J. From, ``Thorvald ii-a modular and re-configurable
  agricultural robot,'' \emph{IFAC-PapersOnLine}, vol.~50, no.~1, pp.
  4588--4593, 2017.

\bibitem{koirala2019deep}
A.~Koirala, K.~Walsh, Z.~Wang, and C.~McCarthy, ``Deep learning for real-time
  fruit detection and orchard fruit load estimation: Benchmarking of
  ‘mangoyolo’,'' \emph{Precision Agriculture}, vol.~20, no.~6, pp.
  1107--1135, 2019.

\bibitem{milioto2018real}
A.~Milioto, P.~Lottes, and C.~Stachniss, ``Real-time semantic segmentation of
  crop and weed for precision agriculture robots leveraging background
  knowledge in cnns,'' in \emph{2018 IEEE international conference on robotics
  and automation (ICRA)}.\hskip 1em plus 0.5em minus 0.4em\relax IEEE, 2018,
  pp. 2229--2235.

\bibitem{zabawa2019detection}
L.~Zabawa, A.~Kicherer, L.~Klingbeil, A.~Milioto, R.~Topfer, H.~Kuhlmann, and
  R.~Roscher, ``Detection of single grapevine berries in images using fully
  convolutional neural networks,'' in \emph{Proceedings of the IEEE Conference
  on Computer Vision and Pattern Recognition Workshops}, 2019, pp. 0--0.

\bibitem{eitel2015multimodal}
A.~Eitel, J.~T. Springenberg, L.~Spinello, M.~Riedmiller, and W.~Burgard,
  ``Multimodal deep learning for robust rgb-d object recognition,'' in
  \emph{2015 IEEE/RSJ International Conference on Intelligent Robots and
  Systems (IROS)}.\hskip 1em plus 0.5em minus 0.4em\relax IEEE, 2015, pp.
  681--687.

\bibitem{valada2019self}
A.~Valada, R.~Mohan, and W.~Burgard, ``Self-supervised model adaptation for
  multimodal semantic segmentation,'' \emph{International Journal of Computer
  Vision}, pp. 1--47, 2019.

\bibitem{mccormac2017semanticfusion}
J.~McCormac, A.~Handa, A.~Davison, and S.~Leutenegger, ``Semanticfusion: Dense
  3d semantic mapping with convolutional neural networks,'' in \emph{2017 IEEE
  International Conference on Robotics and automation (ICRA)}.\hskip 1em plus
  0.5em minus 0.4em\relax IEEE, 2017, pp. 4628--4635.

\bibitem{behley2019semantickitti}
J.~Behley, M.~Garbade, A.~Milioto, J.~Quenzel, S.~Behnke, C.~Stachniss, and
  J.~Gall, ``Semantickitti: A dataset for semantic scene understanding of lidar
  sequences,'' in \emph{Proceedings of the IEEE International Conference on
  Computer Vision}, 2019, pp. 9297--9307.

\bibitem{zaenker2020viewpoint}
T.~Zaenker, C.~Smitt, C.~McCool, and M.~Bennewitz, ``Viewpoint planning for
  fruit size and position estimation,'' \emph{arXiv preprint arXiv:2011.00275},
  2020.

\bibitem{grunnet2018using}
A.~Grunnet-Jepsen, P.~Winer, A.~Takagi, J.~Sweetser, K.~Zhao, T.~Khuong,
  D.~Nie, and J.~Woodfill, ``Using the realsense d4xx depth sensors in
  multi-camera configurations,'' \emph{Intel Corp.}, 2018.

\bibitem{he2017mask}
K.~He, G.~Gkioxari, P.~Doll{\'a}r, and R.~Girshick, ``Mask r-cnn,'' in
  \emph{Proceedings of the IEEE international conference on computer vision},
  2017, pp. 2961--2969.

\bibitem{hartley2003multiple}
R.~Hartley and A.~Zisserman, \emph{Multiple view geometry in computer
  vision}.\hskip 1em plus 0.5em minus 0.4em\relax Cambridge university press,
  2003.

\bibitem{polder2014depth}
G.~Polder and J.~Hofstee, ``Phenotyping large tomato plants in the greenhouse
  usig a 3d light-field camera,'' \emph{American Society of Agricultural and
  Biological Engineers Annual International Meeting 2014, ASABE 2014}, vol.~1,
  pp. 153--159, 01 2014.

\bibitem{halstead2020fruit}
M.~Halstead, S.~Denman, C.~Fookes, and C.~McCool, ``Fruit detection in the
  wild: The impact of varying conditions and cultivar,'' in \emph{2020 Digital
  Image Computing: Techniques and Applications (DICTA)}.\hskip 1em plus 0.5em
  minus 0.4em\relax IEEE, 2020, pp. 1--8.

\bibitem{magistri2020iros}
\BIBentryALTinterwordspacing
F.~Magistri, N.~Chebrolu, and C.~Stachniss, ``{Segmentation-Based 4D
  Registration of Plants Point Clouds for Phenotyping},'' in \emph{IROS}, 2020.
  [Online]. Available:
  \url{http://www.ipb.uni-bonn.de/pdfs/magistri2020iros.pdf}
\BIBentrySTDinterwordspacing

\end{thebibliography}
